\documentclass[preprint,12pt]{elsarticle}

\usepackage[numbers]{natbib}
\usepackage{subfig}

\begin{document}


\title{Mitigating the Impact of Adversarial Attacks in Very Deep Networks}

\author [lb1]{Mohammed Hassanin}
\address[lb1]{School of Engineering and Information Technology, University of New South Wales @ ADFA, Canberra, ACT 2600, Australia. E-mail: m.hassanin@student.unsw.edu.au}

\author [lb2]{Ibrahim Radwan}
\address[lb2]{Faculty of Science and Technology, University of Canberra. E-mail: ibrahim.radwan@canberra.edu.au}

\author [lb15]{Nour Moustafa}
\address[lb15]{School of Engineering and Information Technology, University of New South Wales @ ADFA, Canberra, ACT 2600, Australia. E-mail: nour.moustafa@unsw.edu.au}

\author [lb3]{Murat Tahtali}
\address[lb3]{School of Engineering and Information Technology, University of New South Wales @ ADFA, Canberra, ACT 2600, Australia. E-mail: murat.tahtali@adfa.edu.au}

\author [lb4]{Neeraj Kumar}
\address[lb4]{Thapar Institute of Engineering and Technology, Patiala (Punjab), India. E-mail: neeraj.kumar@thapar.edu}

\cortext[cor1]{Corresponding author: Mohammed Hassanin}

\begin{abstract}
Deep Neural Network (DNN) models have vulnerabilities related to security concerns, with attackers usually employing complex hacking techniques to expose their structures. Data poisoning-enabled perturbation attacks are complex adversarial ones that inject false data into models. They negatively impact the learning process, with no benefit to deeper networks, as they degrade a model’s accuracy and convergence rates. In this paper, we propose an attack-agnostic-based defense method for mitigating their influence. In it, a Defensive Feature Layer (DFL) is integrated with a well-known DNN architecture which assists in neutralizing the effects of illegitimate perturbation samples in the feature space. To boost the robustness and trustworthiness of this method for correctly classifying attacked input samples, we regularize the hidden space of a trained model with a discriminative loss function called Polarized Contrastive Loss (PCL). It improves discrimination among samples in different classes and maintains the resemblance of those in the same class. Also, we integrate a DFL and PCL in a compact model for defending against data poisoning attacks. This method is trained and tested using the CIFAR-10 and MNIST datasets with data poisoning-enabled perturbation attacks, with the experimental results revealing its excellent performance compared with those of recent peer techniques.
\end{abstract}




\maketitle
\section{Introduction}
\label{sec:intro}

\begin{figure*}[h!]
    \centering
       \includegraphics[width=0.95\textwidth]{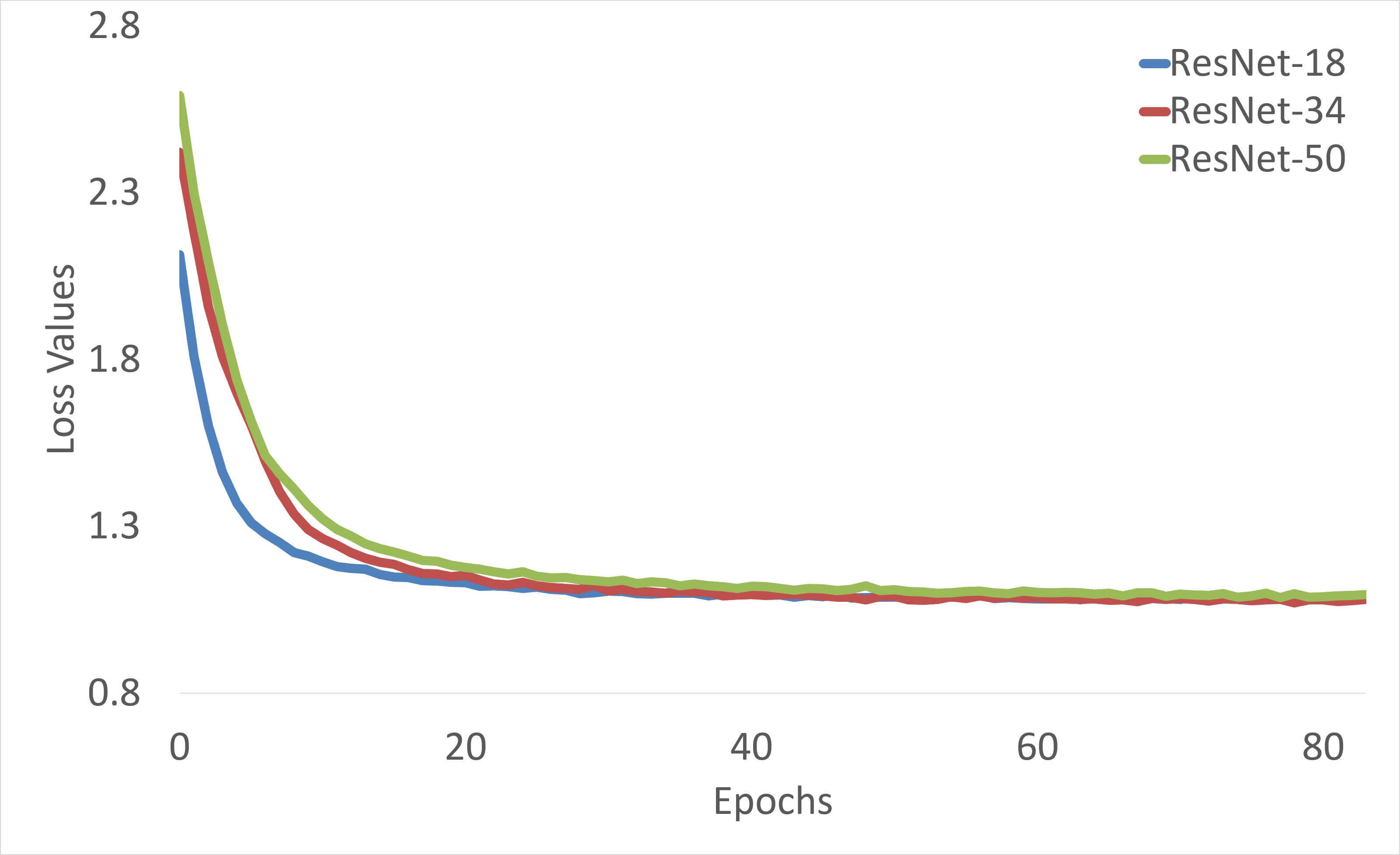}\\(a)\\
       \includegraphics[width=0.95\textwidth]{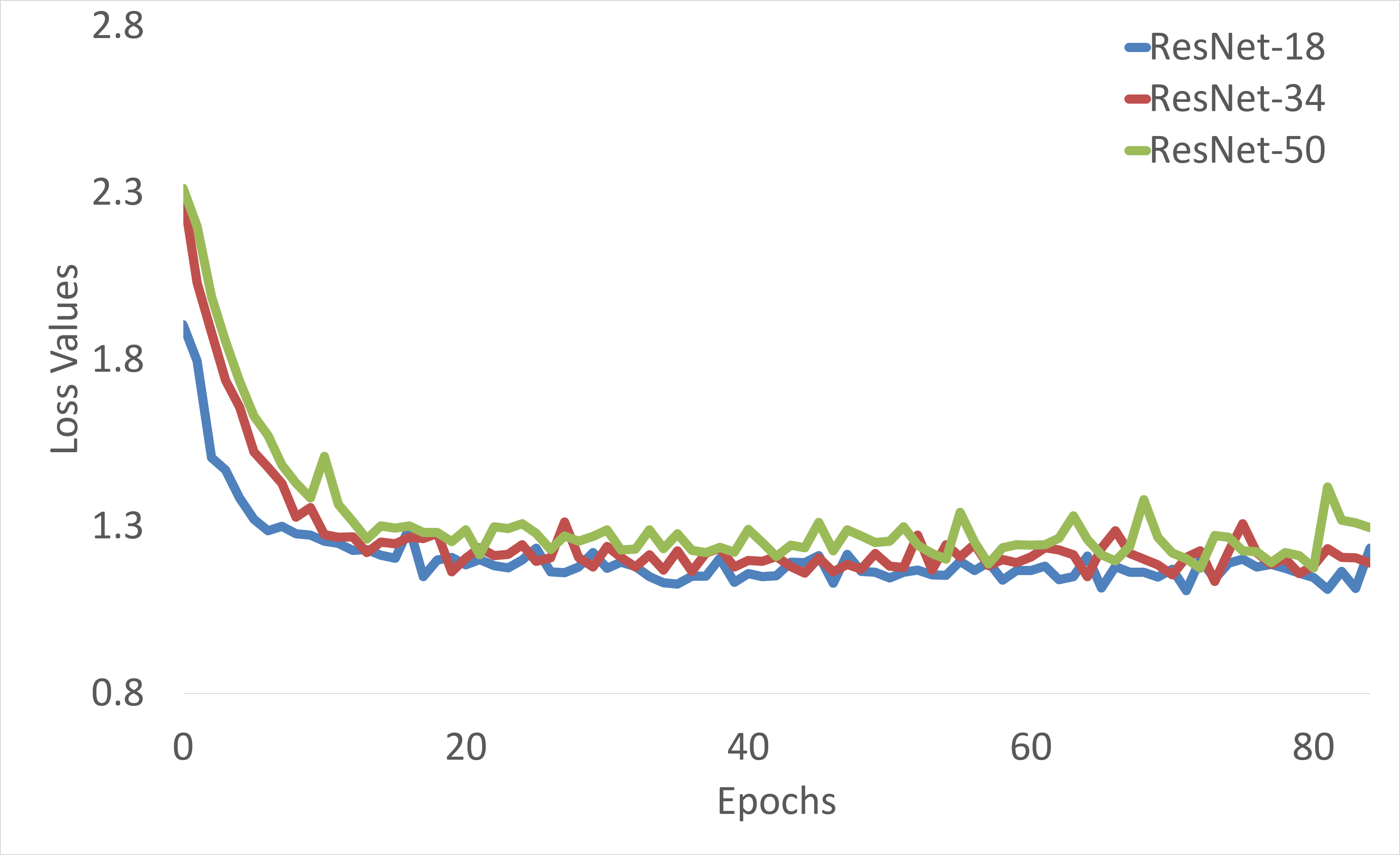}\\(b)
   \caption{The values of training (a) and validation (b) losses of ResNet-18, ResNet-34 and ResNet-50 architectures.}
  \label{fig:motivation} 
\end{figure*}

Deep learning algorithms have been widely used to develop real-world applications, for example, autonomous driving \cite{ackerman2017drive}, bio-metric identification \cite{sanderson2008biometric}, surveillance systems \cite{najafabadi2015deep}, keypoint detection \cite{radwan2019hierarchical}, and cyber-security systems \cite{moustafa2019outlier}. However, these applications are vulnerable to various sophisticated attacking scenarios due to weaknesses in their learning processes. Attackers attempt to compromise the structures of deep learning models, for example, during their stages of training, testing, parameter tuning and creating input datasets, using complex and persistent hacking techniques of which a data poisoning attack is one of the most serious. It makes a model vulnerable and less trustworthy as an attacker illegally alters legitimate data in its training stage to considerably degrade its performance  \cite{moustafa2019outlier, steinhardt2017certified}.

The trustworthiness of deep learning models for combating data poisoning attacks has become one of the main issues for industry and research communities \cite{munoz2017towards}. In order to develop a robust one, adversarial hacking scenarios, such as data poisoning-enabled perturbations, should be designed to estimate to what extent a model correctly classifies data inputs. While such attacks often illegally inject perturbations into input samples, they significantly degrade a model’s performance due to small variations in the feature space between normal and abnormal observations.

To alleviate the negative impacts of attacks, multiple types of defenses in deep learning methods have been introduced in the literature \cite{denoise_1, denoise_2, denoise_3, rifai2011contractive, ball_segedy, bai2017alleviating, luo2015foveation, mustafa2019image}. TThese approaches are classified as three types. Firstly, input transformation-dependent ones eliminate added perturbations and their impacts on data samples and improve a model’s accuracy against attack activities. Secondly, network modification-based ones adapt a network’s architecture, such as by adding parallel branches to its baseline, to protect it against adversarial attacks  \cite{gao1702masking, cisse2017houdini}. Thirdly, gradient mask-based ones attempt to increase the linearity of the training process to reduce adversarial effects in a less risky manner \cite{papernot2016limitations, salman_adversarial, lyu2015unified, nguyen2018learning, advers_training_3, kurakin2016adversarial}.

Existing defense approaches are capable of enhancing a model’s efficiency for handling some attack types rather than others. However, input transformation-based ones require the perturbation scale to be minimal while regularization- and network modification-based ones are more vulnerable to white-box attacks, such as those of data poisoning. In this paper, we propose a generative adversarial defensive method with a regularized feature layer which considers the risk of its performance for a very deep network degrading. 

Moreover, we observe that, by training using adversarial settings, the problem of a gradient vanishing is more likely to occur in a deep architecture than small network, even with a small adversarial perturbation budget, that is, a shallow network’s convergence is slightly better than that of a deeper one. The issue of decreasing accuracy when stacking more layers together in a network is addressed in \cite{he2016deep} but, given adversarial attacks, is highlighted again for deeper networks. This means that, even with a slight perturbation applied to a benign input instance, the accuracy of classification degrades drastically, particularly with deeper networks. This observation is validated by training three
ResNet architectures(\textit{i.e.} ResNet-18, ResNet-34 and ResNet-50) using the same training settings(\textit{e.g.} adversarial PGD attack for $10$ iterations), with the
loss curves depicted in Figure \ref{fig:motivation}. The deeper the architecture, the less convergence of the loss values as well as a greater decrease in accuracy.

To address the degradation of accuracy in a deep architecture while training using adversarial settings,  \textit{e.g.}, data poisoning, we propose augmenting each ResNet architecture with a Defensive Feature Layer (DFL), which neutralizes the impact of the perturbations added in the feature space, and a regularization term that improves the decision boundaries between the learned classes. The proposed DFL plays a crucial role in denoising the perturbations in the feature space and produces defensive feature maps which address the accuracy issue. Although the DFL added to defend against adversarial attacks is solid, it represents only one type of defense (\textit{i.e.} network modification). To increase the robustness of the proposed method against data poisoning attacks, we regularize the extracted features in the hidden space using a discriminative loss function called Polarized Contrastive Loss (PCL). This increases the separation between samples in different classes in the feature space and maintains the similarity of those in the same one. The proposed method is tested on the two common datasets CIFAR-10 \cite{cifar_10} and MNIST \cite{MNIST} and shows significant improvements compared with baseline methods.

The key contributions of proposed method are summarised as follows:
\begin{itemize}
    \item A new generative defensive method mitigates the impacts of adversarial attacks and addresses the issue of accuracy degradation and model convergence in Deep Neural Networks (DNNs).
    \item A new regularization function efficiently maps the feature space to the output space to enable better classification accuracy.
    \item The two components of the proposed method are combined to form a robust defense against adversarial and data poisoning attacks, with the experimental results obtained using two common datasets presented.
\end{itemize}
\section{Related Work}
\subsection{Adversarial Deep Learning}

Deep learning models have various vulnerabilities related to security concerns \cite{haq2017advanced, sagduyu2019adversarial} which can involve the exposure of their data inputs, structures, training and testing phases, parameter tuning processes, feature extraction techniques or outputs. Attackers often use complex hacking techniques called Advanced Persistence Threats (APTs) \cite{haq2017advanced} to breach the security principles of Confidentiality, Integrity and Availability (CIA) and the potential designs of deep learning algorithms and their inputs \cite{sagduyu2019adversarial}. Attacks on confidentiality attempt to infiltrate a model’s structure, including its training and testing phases and parameter tuning process, in order to expose its datasets. Integrity attacks aim to alter, and inject malicious activity into, a model’s training and testing sets. Attacks on availability try to prevent legitimate users from accessing a model’s processes for extracting features and inferring suspicious activities \cite{moustafa2019outlier, dreossi2018semantic}.

To study the robustness and trustworthiness of developing algorithms while exploiting them using hacking scenarios such as data poisoning, adversarial deep learning algorithms have emerged \cite{sagduyu2019adversarial,dreossi2018semantic}. Adversaries try to discover weak stochastic properties and dynamic data distributions in deep learning models \cite{moustafa2019outlier}. Examples of adversarial attacks with perturbation functions confuse a model’s classifiers and degrade its performance. The idea of adversarial examples was used to develop the L-BFGS model that could improve its accuracy \cite{ball_segedy}.

Using a Fast Gradient Sign Method (FGSM), Goodfellow et al. \cite{FGSM} generated adversarial examples $\tilde{x}$ from a normal sample $x$ by linearizing the model’s maximization. The potential process of FGSM is to move in the opposite direction of the gradient of a loss function.

The Basic Iterative Method (BIM) was proposed and integrated with FGSM to generate adversarial examples by repeating it for multiple iterations. At every iteration, the generated values were clipped to represent a small change change from those of the original samples \cite{BIM}.

Similar to BIM \cite{BIM}, Yinpeng et  \cite{MIM} proposed the Momentum Iterative Method (MIM) as another variant of the FGSM attack that generates adversarial examples through the iterative process but defined a new term, \textit{decay} $\mu$ parameter, to stabilize the direction of the gradient.

Madry et al. \cite{PGD} showed that the BIM \cite{BIM} is a projected descent for negative loss function. Based on that, they developed Projected Gradient Descent (PGD) to address this issue through focusing to select some points around the clean input sample within the $L_\infty$ to address this issue. It has proven to be a strong attack and its adversarial training is relatively more robust against all attacks.
%
\subsection{Defense Approaches}
Traditional machine learning models have been proven to be vulnerable to complex attacks. However, they can be defensive and robust using various types of techniques, such as adversarial training  \cite{advers_training_1, advers_training_2, advers_training_3, advers_training_4}, network architecture modification \cite{network_3, network_1, network_2} and input transformation \cite{input_1, input_2, input_3, input_4, denoise_1, denoise_2, denoise_3}.

Adversarial training, which mainly trains a whole model based on injected adversarial examples, is the most popular defense against attacks. It can be considered as a noise added to the training process to improve the robustness of the model. However, it requires more time than the normal training process to generate and inject perturbed clean ones and more space to accommodate its additional steps.

Another way of improving the robustness of a machine learning method is to transform its input samples to eliminate the effect of an attack before feeding it into the model. Autoencoders have been used to remove the noise in an input sample and reduce the effect of an attack \cite{gu2014towards}.  Osadchy \textit{et. al.,}  \cite{osadchy2017no} processed an image using a set of filters to remove the noise. In \cite{das2017keeping}, Das \textit{et. al.,} used JPEG compression as a pre-processing step to eliminate the noise from an adversarial example. Liao et al. employed the reconstruction error of high-level features to guide de-noisers \cite{denoise_2}. Xie \textit{et. al.,} alleviated the adversarial effects by pre-processing input samples using various techniques such as random re-sizing and padding.

Another set of defenses is based on making the classifiers more defensive against perturbations  \cite{papernot2017practical, papernot2016towards, miyato2015distributional}, with these methods dependent on sharpening the decision boundaries of Cross-entropy (CE) for better classification. In \cite{ross2018improving}, Ross \textit{et. al.,} penalized a change in the output due to perturbations by regularizing the gradient of the classifier’s loss (CE). Nayebi \textit{et. al.,} used non-linear activations to increase the robustness of a network and proposed a loss function that encouraged similar activations to be in their regime. Although these methods have shown their robustness against white-box attacks, they are vulnerable to black-box ones \cite{advers_training_3}. 

As each of the previous methods focuses on only one style of defense (\textit{i.e.,} network modification, gradient mask or input transformation), their algorithms fail in some way one side of the attacks. Based on that, our proposed approach combines two main defenses: 1) network modification, namely DFL, to increase robustness against black-box attacks; and 2) the proposed PCL as a regularizer added to the CE to discriminate among classes in the feature space and increase the linearity of the training process. In this way, it defends against all types of attack.

\section{Proposed Method}
\begin{figure*}
    \includegraphics[width=0.99\textwidth]{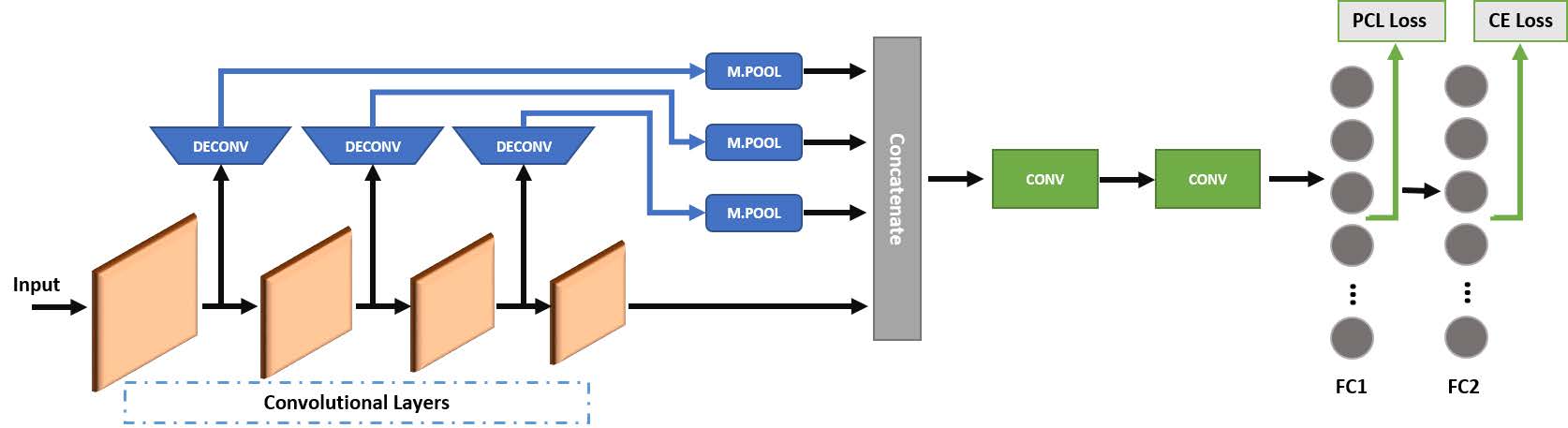}
    \caption{Pipeline of the proposed adversarial training: The activation maps of each convolutional layer is sent to a deconvolutional block, and then their outputs are concatenated to form the defensive features. The Polarised Contrastive Loss (PCL) is used to regularise the hidden space of the network, while the final output is regularised with the Cross Entropy (CE) loss function.}
    \label{fig:pipeline}
\end{figure*}

\subsection{Method Overview}

Training a Classifier $f_\theta: X \rightarrow Y$, where $\theta$ represents the learned parameters, encodes a compact form of the relationship between input and output spaces. The input space involves the training instances (\textit{i.e.,} images), $X =\{x_1, x_2, ..., x_n\}$, where $x_i$ is an input instance and $n$ is the total number of instances in the input space while the latter represents the set of possible classes, that is, $Y=\{y_1, y_2, ..., y_k\}$, where $y_j \in \mathcal{R}$ represents the $j^{th}$ class in the output space and $k$ is the total number of~classes. 

To mitigate the impact of adversarial attacks, we augment the input space with adversarial examples generated
by adding perturbations to input instances
, with $\tilde{x} = x_i + \epsilon$  an adversarial example created from a benign instance, $x_i$, and $\epsilon$ the perturbation scale of an adversarial attack. These examples are generally designed to increase the robustness and improve the generalization capacities of the classifiers. In our method, the input space is obfuscated by different types of attacks to produce various adversarial examples.

Adding adversarial examples while training a classifier presents two challenges. Firstly, the features extracted from the input instances and passed to the classification step need to be highly defensive against different types of adversarial attacks. This enables the features of benign and adversarial instances, which belong to the same class, to be close to each other in the feature space. Secondly, the classification boundaries between different classes are required to be separable. The proposed method is designed to address these two challenges. A DFL is added to a DNN to neutralize the impacts of illegitimate perturbations in the feature space. Then, the PCL is introduced to enhance inter-class variabilities and discrepancies among different classes as well as improve intra-class compactness. The pipeline of the proposed method is depicted in
 Figure \ref{fig:pipeline} 
%
\subsection{Defensive Feature Layer}
The baseline DNN ResNet \cite{he2016deep} comprises a sequence of convolutional layers, each of which is a mix of convolution, pooling, normalization, skip connections and non-linear activation functions, with the output of each passed as an input to the next layer. Plugging adversarial examples into such deep networks during training results in degraded accuracy, to avoid which we add a DFL while extracting the features of the input instances. This layer tries to remove the impact of the added noise (\textit{i.e.,} perturbations) in the feature space and consists of a concatenation of the output from the convolutional layers and sequence of deconvolutional blocks.

\begin{figure*}
    \centering
       \includegraphics[width=0.9\linewidth]{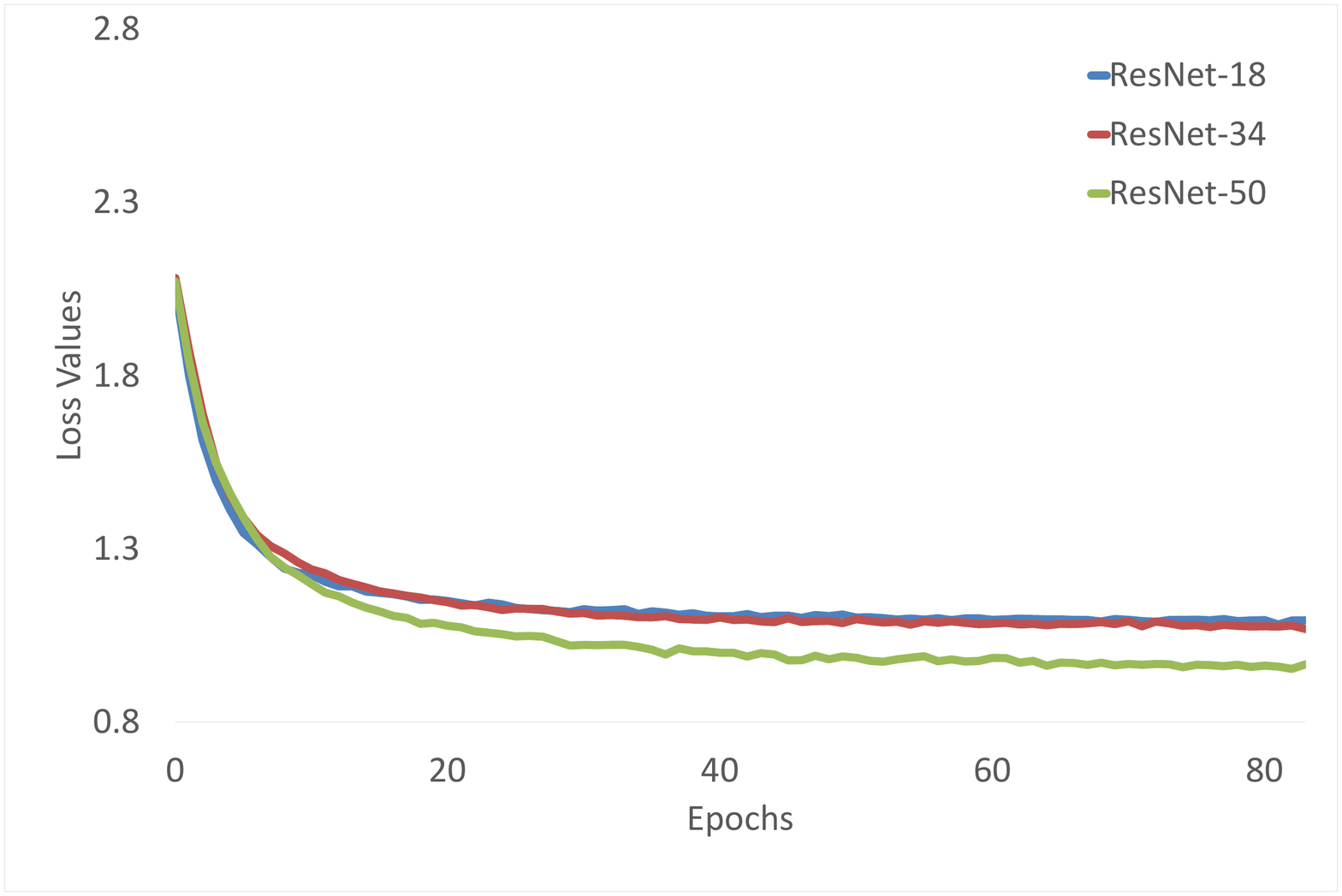}\\(a)\\
           \includegraphics[width=0.9\linewidth]{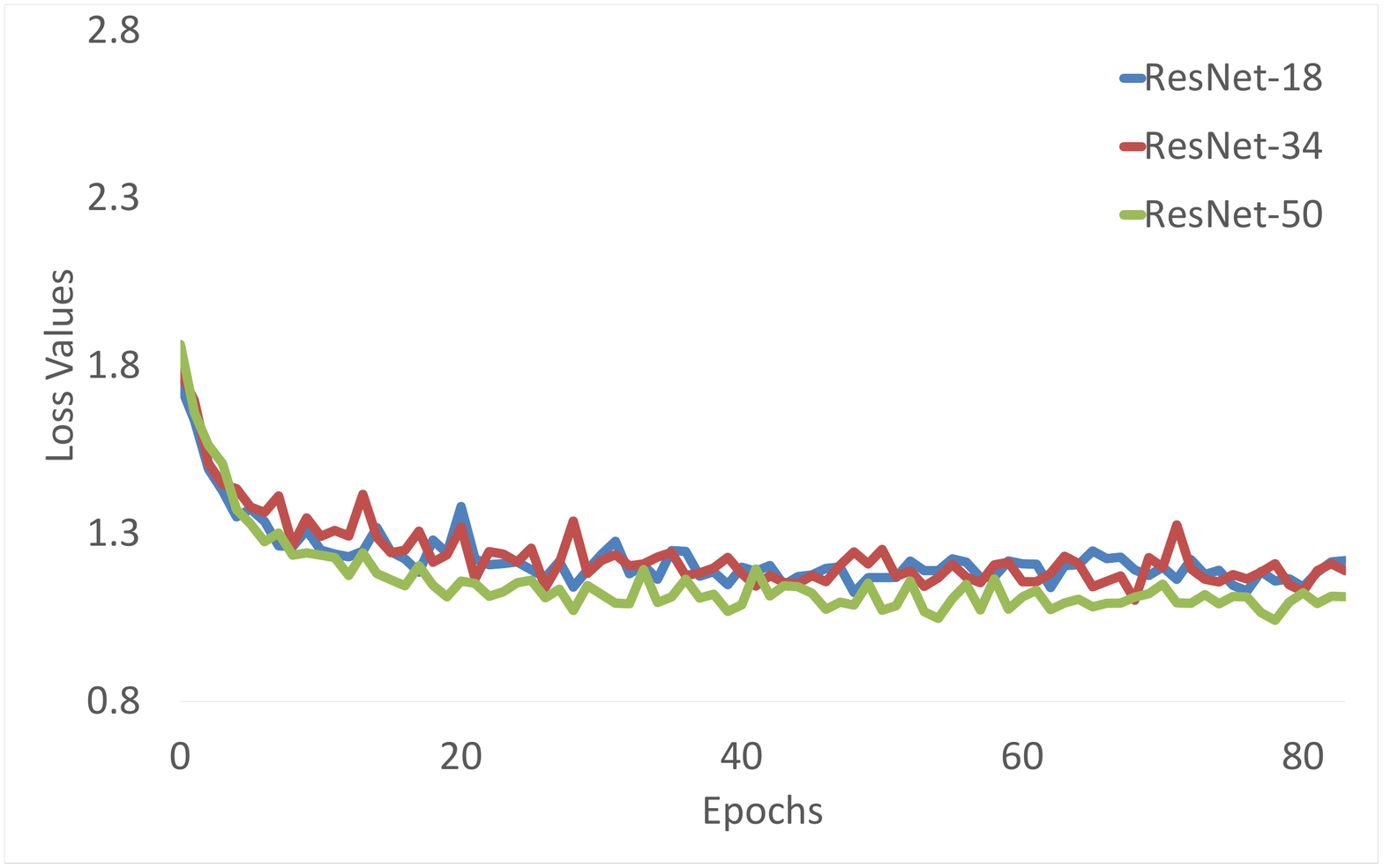}\\(b)
   \caption{Error values of training (a) and validation (b) deep architectures (ResNet-18, ResNet-34 and ResNet-50) under the PGD attack.
}
  \label{fig:dfl_losses} 
\end{figure*}

\paragraph{Deconvolution Blocks:}
As the convolutional layers contain sequentially connected convolution and pooling operations, the deconvolutional blocks may guide the network to focus on extracting the salient features from its input instances but image details may be lost. Although these blocks, especially on the shallow layers, have the capacity to compensate for any loss, plugging them into only the output from the last convolutional layer (\textit{i.e.,} as a decoder \cite{mao2016image}) will not be a good option for neutralizing the impact of adversarial attacks as image details will already be lost. In our network, we solve this issue by adding a deconvolutional block after each convolutional layer.
The deconvolutional blocks learn kernels that neutralize the added perturbations and retain image details because they make top–down projections by mapping the activation maps back to the input space. The pooling layer is used to unify the feature maps of all the layers. Technically, the activation maps  $F_i \in \mathcal{R}^{c\times w\times h}$ are generated from the convolutional layers, where $c$ is the number of learned kernels in the convolutional layer, while $w$ and $h$ are the width and heights of the extracted activation maps, respectively. Each of these maps is passed to a deconvolutional layer, as shown on the upper branch of Figure \ref{fig:pipeline}), which helps to reduce perturbations in the adversarial samples and obtain the extracted maps that are similar to the corresponding ones of the clean samples in the feature space.

The output from each deconvolutional layer, $D(F_i)$, is concatenated with the output of last convolutional layer, $F_m$, where $D$ represents the internal operations of a deconvolutional block; whereas $m$ is the number of the convolutional layers in the baseline network.  The features extracted from the DFL are
\begin{equation}
    F = F_m \oplus \mathbin\Vert_{j=1}^{m-1} D(F_j),
\end{equation}
where $\mathbin\Vert$ refers to the concatenation of the maps, which are extracted from the deconvolutional blocks and $\oplus$ that of the outputs from the deconvolutional block and final convolutional layer in the network. Also, these feature maps are passed to max-pooling operations to satisfy dimensional requirements. The skip connections between the deconvolutional blocks and concatenation layer require the model to be aware of the context of the extracted features as well as enable back-propagation of the gradients to the shallow layers which yields a better convergence while training the model and prevents the gradient vanishing. Then, the concatenated features become more defensive against adversarial attacks which leads to the training and validation error rates decreasing more for ResNet-50 than for ResNet-18 and ResNet-34, as shown in Figure \ref{fig:dfl_losses}, contrary to what can be observed in  Figure \ref{fig:motivation}. This demonstrates the effectiveness of using the DFL to defend against adversarial attacks in very deep architectures.

\subsection{Polarised Contrastive Loss}
\begin{figure*}
    \centering
    \subfloat[a]{
      \includegraphics[width=0.2\textwidth]{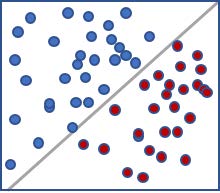}}
  \subfloat[b]{
      \includegraphics[width=0.2\textwidth]{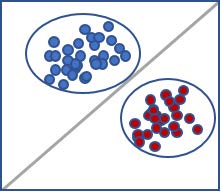}}
	   \subfloat[c]{
      \includegraphics[width=0.2\textwidth]{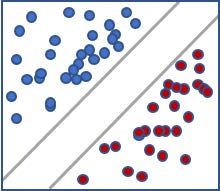}}
	   \subfloat[d]{
      \includegraphics[width=0.2\textwidth]{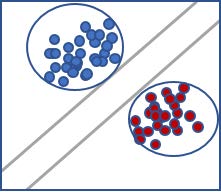}}
  \caption{A visual representation of the decision boundary btween two classes using: a) Cross Entropy loss, b) Center loss \cite{center_loss}, c) Additive Margin with Cross Entropy loss \cite{arcface}, and the Polarised Contrastive Loss (PCL) functions.}
  \label{fig:pcl} 
\end{figure*}
A DNN for multi-class classification is trained by passing the input samples through different layers and blocks of the network. The extracted features are then mapped to the output space by an objective function learning this mapping between the feature and output spaces, with CE the traditional one defined as
\begin{equation}
\label{eq:CE}
    \mathcal{L}_{CE}(f_i, y) = \frac{1}{n}\sum_{i = 1}^{n} -\log \frac{\exp(\phi_i)}{\sum_{j} \exp(\phi_j)},
\end{equation}
where, $f_i \in R^d$ represents the extracted features, with $d$ is the size of the extracted features. 

Also, $\phi_{i} = f^T_i\cdot w_{y_{i}}$ is the dot product between the extracted features of a sample $x_i$ and the corresponding vector representing the true class, $w_{y_{i}}$. In adversarial training, CE usually fails to discriminate between the benign and adversarial samples in the feature space because it does not explicitly enforce a robust margin between learned classes. This leads to the features of samples obtained from different classes overlapping which makes it easier for an attacker to deceive the classifier even if the perturbation scale is minimal. To address this issue and improve the robustness of the trained classifier in an adversarial setting, two constraints need to be enforced when mapping the feature space to the output one. Firstly, the intra-class variability should be minimized and, secondly, the inter-class discrepancy maximized. 

Inspired by \cite{center_loss}, and to satisfy the first constraint, we force the features of input samples from the same class to be assembled around the center of their class’s weight, $w_{y_{i}}^{c}$. This is performed by computing the distances between these features and the learned centroids of their corresponding classes as
\begin{equation}
\label{eq:C}
    \mathcal{L}_{intra}(f_i, y) = \sum_{i} \parallel f_i - w_{y_{i}}^{c} \parallel
\end{equation}

To comply with the second constraint, the inter-class variability considers two parts: 1) maximizing the difference between the features of one class $y_i$ and centroids of the others; and 2) maximizing the discrepancy between the centroids of different classes. These two quantities represent the inter-regularization term in our objective function which is computed as
\begin{equation}
\label{eq:Arc}
    \mathcal{L}_{inter}(f_i, y) = \frac{1}{k-1}\sum_{j\neq y_i}^{K-1}\parallel f_i - w^{c}_j\parallel + \arccos(w_{y_i}^c\ w^c_j),
\end{equation}
where, the first term, $\parallel f_i - w^{c}_j\parallel$, encourages the features of class $y_i$ to be separated from the other classes' centroids. The second term, $\arccos(w_{y_i}^c\ w^c_j)$ enables the decision margins between the centroids of the classes to be separated sufficiently to prevent the overlapping of features from different classes (more details are provided in \cite{arcface}). Visual representations of the CE and added regularization terms are depicted in Figure \ref{fig:pcl}. 

The objective function (\textit{i.e.,} the PCL) for training the proposed network consists of the CE and regularization terms as

\begin{equation}
    \label{eq:objective}
    \mathcal{L}(f_i, y) = \mathcal{L}_{CE} + \mathcal{L}_{inter} + \mathcal{L}_{intra}
\end{equation}
The proposed method combines the advantages of the DFL, which neutralizes the impact of the added perturbations, and using the PCL as an objective function to reduce the overlap of samples of different classes in the feature space. This helps to nullify the impacts of adversarial attacks, enforces robust decision boundaries between the classes and centralizes the features of samples from the same category around the centroids of their classes.  
\section{Experiments}
In this section, details of the implementation of the proposed method are provided and quantitative comparisons of it and baseline approaches conducted to validate its strength. The performances of the complete proposed method (with both the DFL and PCL) are discussed and then the effectiveness of these components considered separately in more detail to demonstrate their impacts in the adversarial training settings. Generally, the proposed method outperforms state-of-the-art techniques and introduces a new benchmark for adversarial training on the \cite{cifar_10} and MNIST \cite{MNIST} datasets.


\begin{table}

\caption{Comparison between the proposed method (PCL + DFL) and the baseline approaches for MNIST dataset. \label{table:All_mnists}}

\centering
\begin{tabular}{|l|ccc|}
\hline 
\multicolumn{4}{|c|}{\textbf{No Defense}}\tabularnewline
\hline 
Defenses &  FGSM & PGD-10 & C\&W\tabularnewline
\hline 
ResNet-50 & 4.9 & 0.0 & 0.2\tabularnewline
\hline 
Mustafa et al. \cite{salman_adversarial}  & 31.1 & 19.9 & 29.1\tabularnewline
\hline 
Dynamic \cite{dynamic} & 14.04 & 0.0 & 0.0 \tabularnewline
\hline 
Proposed Method & \textbf{89.93} & \textbf{57.54} & \textbf{84.14}\tabularnewline
\hline
\multicolumn{4}{|c|}{\textbf{Adversarial Training}}\tabularnewline
\hline 
Defenses & FGSM & PGD-10 & C\&W \tabularnewline
\hline 
ResNet-50 & 59.50 & 57.19 & 57.09\tabularnewline
\hline 
Mustafa et al. \cite{salman_adversarial}  &  53.10 & 34.50 & 40.90\tabularnewline
\hline 
Dynamic \cite{dynamic} &  95.34 & 91.63 & 91.47\tabularnewline
\hline 
Proposed Method &  \textbf{98.68} & \textbf{98.31} & \textbf{98.18}\tabularnewline
\hline 

\end{tabular}

\end{table}

\subsection{Implementation Details}
The ResNet-50 architecture is used as the core of the proposed method, with both the DFL and PCL examined while embedded in it. In the experiments, we conduct adversarial training using three attacks: the  FGSM \cite{FGSM}, PGD \cite{PGD} and C\&W \cite{candw}. The parameters of the architecture are the default ones of ResNet-50 and the kernel size of each deconvolutional block  $1\times1$, with the max-pooling layers adapted to make the size of each feature map the same as that of the last convolutional layer of the network. The number of training epochs is $100$ and $50$ for the CIFAR-10 and MNIST datasets, respectively, with the learning rate $0.1$ and batch size $64$ in all the experiments. The number of iterations of an adversarial attack for poisoning a sample is $10$, with the perturbation scale $0.03$ for the CIFAR-10 and $0.3$ for the MNIST datasets while, for the C\&W parameters, c = $0.1$.

\subsection{Baselines} The proposed method is compared with three baseline approaches: 1) the ResNet-50 architecture with a CE loss; 2) a gradient mask-based technique which is the closest to our proposed objective function \cite{salman_adversarial}; and 3) a dynamic training strategy  \cite{dynamic}. The comparison is performed with no attack provided in the training process (\textit{i.e.,} 'no defense') but some in the testing one, and then with attacks to poison the input samples in the training and testing phases (\textit{i.e.,} adversarial training).  
\subsection{Results and Discussion}
\textbf{MNIST:} The results obtained by the proposed and three baseline methods on the MNIST dataset presented in Table \ref{table:All_mnists} show that the former outperforms the others by more than $60\%$ in the ’no defense’ settings and by $37\%$ in the adversarial ones. This introduces a new benchmark for the MNIST dataset with adversarial training and indicates that ’no defense’ training is more difficult than black-box defensive methods since the trained model does not have any security which proves that the proposed method is more robust against both white- and black-box attacks.
%

\begin{table}
\caption{Comparison between the proposed method (PCL + DFL) and the baseline approaches for CIFAR-10 dataset. \label{table:All_cifar}}
\centering
\begin{tabular}{|l|ccc|}
\hline 
\multicolumn{4}{|c|}{\textbf{No Defense}}\tabularnewline
\hline 
Defenses & FGSM & PGD-10 & C\&W\tabularnewline
\hline 
ResNet-50 & 21.4  & 0.01 & 0.6\tabularnewline
\hline 
Mustafa et al. \cite{salman_adversarial} & \textbf{67.70} & 27.20 & 37.30\tabularnewline
\hline 
Dynamic \cite{dynamic} & 2.20 & 0.0 & 0.0\tabularnewline
\hline 
Proposed Method & 43.85 & \textbf{38.70} & \textbf{53.07} \tabularnewline
\hline 
\multicolumn{4}{|c|}{\textbf{Adversarial Training}}\tabularnewline
\hline
Defenses & FGSM & PGD-10 & C\&W\tabularnewline
\hline
  ResNet-50 &  2.2 & 0.0 & 0.0\tabularnewline
  \hline
 Mustafa et al. \cite{salman_adversarial} & 75.80 & 46.70 & 51.80\tabularnewline
 \hline 
Dynamic \cite{dynamic}& 52.81 & 48.06 & 37.26\tabularnewline
\hline 
Proposed Method & \textbf{91.80} & \textbf{74.50} & \textbf{61.30}\tabularnewline
\hline 
\end{tabular}
\end{table}


\noindent\textbf{CIFAR-10:} The performances of the proposed method using the CIFAR-10 dataset depicted in Table \ref{table:All_cifar} show they are better than those of the three baseline approaches by more than $10\%$ with ’no defense’ settings for both the PGD-10 and C\&W attacks whereas  \cite{salman_adversarial} achieves better results for the FGSM attack. Nevertheless, the proposed method is superior to all the baseline approaches by  $10$  $28\%$ with adversarial settings for the different types of attacks. Therefore, as the proposed method achieves significant results for both the ’no defense’ and adversarial training settings, it represents the benchmarks for both types of attack. Finally, the results reported in Tables \ref{table:All_mnists} and \ref{table:All_cifar} validate the robustness of the proposed method against different types of adversarial attacks.
\subsection{Ablation Analysis}
\begin{figure*}
    \centering
      \includegraphics[width=0.9\linewidth]{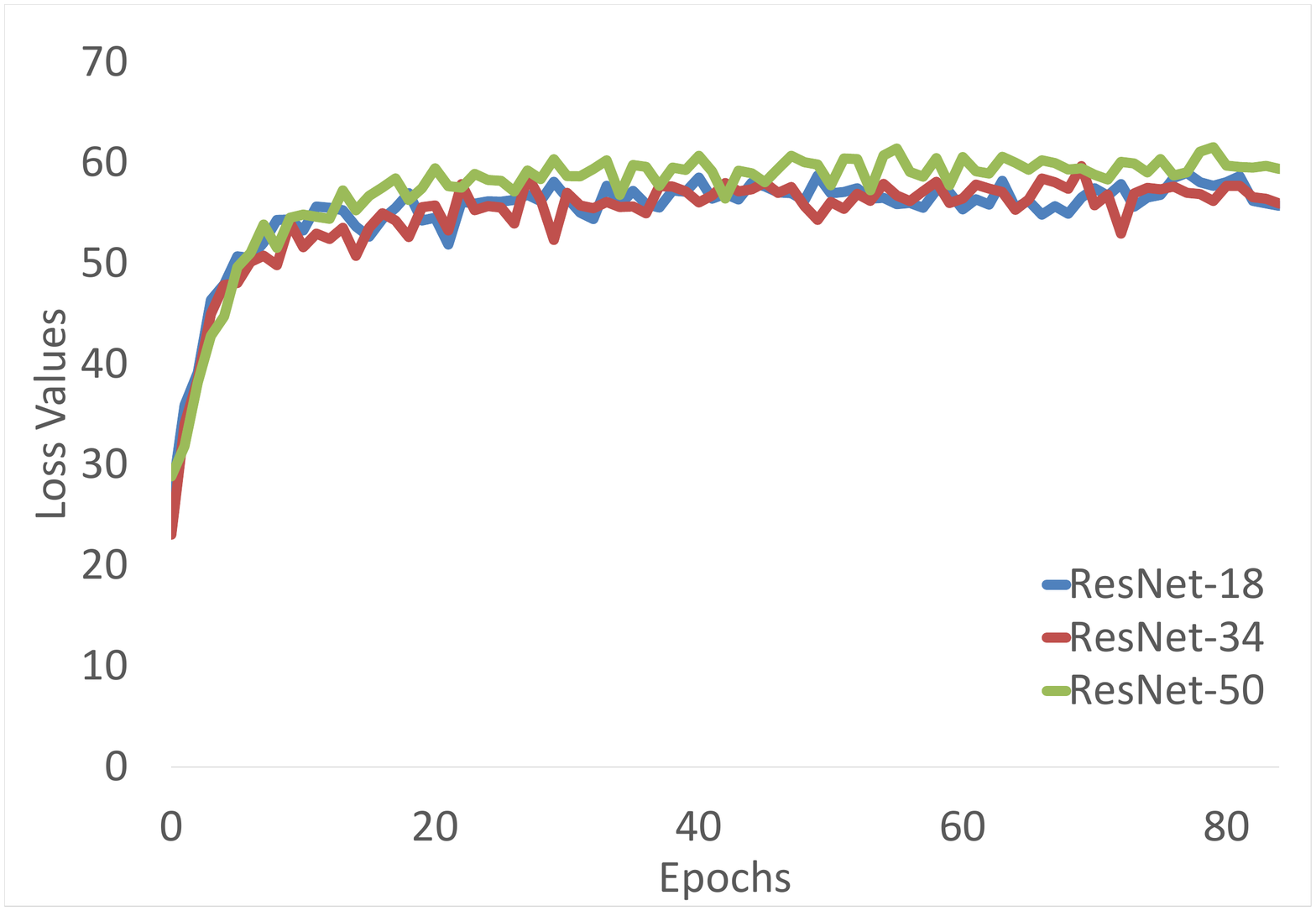}\\(a)\\
      \includegraphics[width=0.9\linewidth]{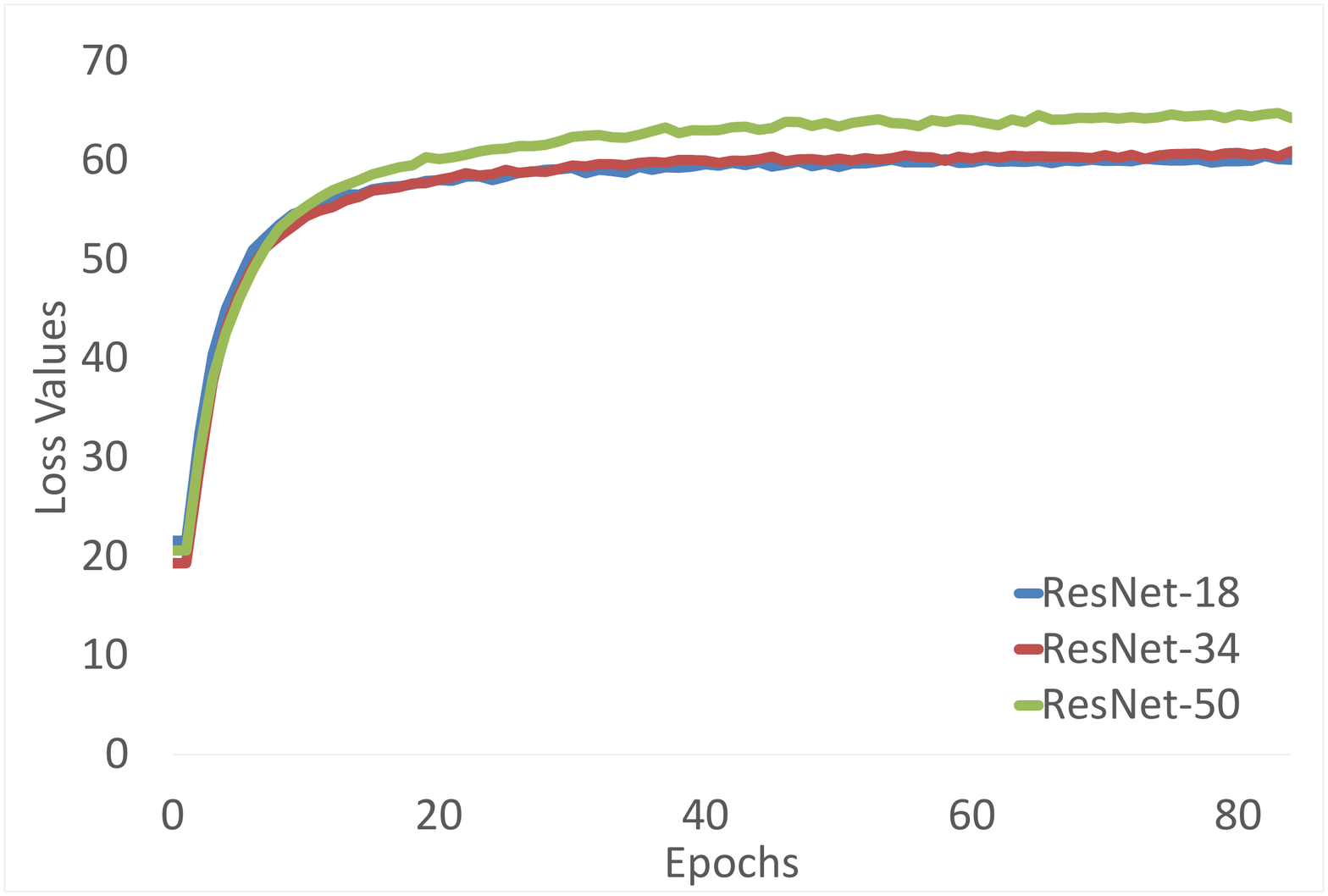}\\(b)
   \caption{Three network architectures (ResNet-18, ResNet-34 and ResNet-50) have been merged with the proposed DFL, to tackle the problem of deep learning in adversarial training. The validation accuracy (a), and training accuracy (b) illustrate the overall impact of using DFL on the deep networks.}
  \label{fig:dfl_accuracy} 
  \vspace{-0.5em}
\end{figure*}

\textbf{Effectiveness of Defensive Feature Layer:}
The impact of the DFL on the whole training process is investigated. Initially, \ref{fig:dfl_accuracy} shows the validation and training accuracy values using the DFL embedded in the different ResNet architectures under one of the most difficult attacks (\textit{i.e.,} PGD). As mentioned in Section  \ref{sec:intro}, adversarial attacks eliminate the power of very deep networks. Therefore, the entire theory of deep networks is not valid in the presence of adversarial attacks. To justify how the DFL addresses this issue, three network architectures, ResNet-18, ResNet-34 and ResNet-50, are trained using a PGD attack. In Figures \ref{fig:motivation}, \ref{fig:dfl_losses} and \ref{fig:dfl_accuracy}, it is very clear that the deeper the model, the better the results obtained when the DFL is embedded in the architectures. Then, the training curve of ResNet-34 shows higher accuracy than that of ResNet-18 whereas ResNet-50 outperforms both, with a significant difference of $7.1\%$ (from $60.5\%$ to $64.8\%$), while the training error reduces to $12\%$.
\paragraph{MNIST:} 
The results obtained from embedding a DFL in each ResNet architecture compared with those from the three baseline approaches using the MNIST dataset are presented in Table \ref{table:DFL_mnist}. The proposed method outperforms the others with more than $60\%$ with the ’no defense’ settings which validates its robustness, and is better than them by $\sim 3 - 7\%$ with the adversarial ones.

\begin{table}
\caption{Comparison between the proposed method (using DFL only) and the baseline approaches for MNIST dataset.
\label{table:DFL_mnist}}
\centering
\begin{tabular}{|l|ccc|}
\hline 
\multicolumn{4}{|c|}{\textbf{No Defense}}\tabularnewline
\hline 
Defenses & FGSM & PGD-10 & C\&W \tabularnewline
\hline 
ResNet-50 & 4.9  & 0.0 & 0.2 \tabularnewline
\hline 
Mustafa et al. \cite{salman_adversarial} & 31.1 & 19.9 & 29.1\tabularnewline
\hline 
Dynamic \cite{dynamic} & 17.04 & 0.0 & 0.0  \tabularnewline
\hline 
Proposed Method (DFL) & \textbf{93.08} & \textbf{59.60} & \textbf{87.0} \tabularnewline
\hline 
\multicolumn{4}{|c|}{\textbf{Adversarial Training}}\tabularnewline
\hline 

Defenses & FGSM & PGD-10 & C\&W\tabularnewline
\hline 

ResNet-50   & 59.50 & 57.19 & 57.09 \tabularnewline
\hline 

Mustafa et al. \cite{salman_adversarial} &   53.1 & 34.50 & 40.90 \tabularnewline
\hline 

Dynamic \cite{dynamic} & 95.34 & 91.63 & 91.47\tabularnewline
\hline 

Proposed Method (DFL) & \textbf{97.30} & \textbf{96.63} & \textbf{97.13}\tabularnewline
\hline 

\end{tabular}
\end{table}

\paragraph{CIFAR-10:} 
The performances of the proposed method using the CIFAR-10 dataset presented in Table \ref{table:DFL_cifar} show that it outperforms the three baseline methods with more than $10\%$ with the ’no defense’ settings for both the PGD-10 and C\&W attacks while the method in \cite{salman_adversarial} achieves better results for the FGSM one. However, the proposed method is better than all the others by $\sim 10-28\%$ with the adversarial settings for the different types of attacks. Overall, the proposed method achieves significant results for both the ’no defense’ and adversarial training settings.

\begin{table}
\caption{Comparison between the proposed method (using DFL only) and the baseline approaches for CIFAR-10 dataset.
\label{table:DFL_cifar}}
\centering
\begin{tabular}{|l|ccc|}
\hline 
\multicolumn{4}{|c|}{\textbf{No Defense}}\tabularnewline
\hline 
Defenses &  FGSM & PGD-10 & C\&W\tabularnewline
\hline 
ResNet-50 & 21.4  & 0.01 & 0.6 \tabularnewline
\hline 
Mustafa et al. \cite{salman_adversarial} & \textbf{67.70} & 27.20 & 37.30\tabularnewline
\hline 
Dynamic \cite{dynamic} & 2.20 & 0.0 & 0.0\tabularnewline
\hline 
Proposed Method (DFL)& 44.60 & \textbf{35.40} & \textbf{48.50}\tabularnewline
\hline 
\multicolumn{4}{|c|}{\textbf{Adversarial Training}}\tabularnewline
\hline 
Defenses & FGSM & PGD-10 & C\&W \tabularnewline
\hline 
ResNet-50& 2.2 & 0.0 & 0.0\tabularnewline
\hline 
Mustafa et al. \cite{salman_adversarial}& 75.80 & 46.70 & 51.80\tabularnewline
\hline 
Dynamic \cite{dynamic} &52.81 & 48.06 & 37.26\tabularnewline
\hline 
Proposed Method (DFL)& \textbf{90.31} & \textbf{64.67} & \textbf{57.30} \tabularnewline
\hline
\end{tabular}
\end{table}
In summary, securing against black-box attacks is considered less dangerous than a ’no defense’ strategy, we claim that this type of modification addresses the four issues of a DNN’s loss of convergence, accuracy degradation, black-box attacks and securing ’no-defense’ training. Also, both white- and black-box attacks are expensive in terms of time and space. 

\noindent\textbf{Effectiveness of Polarised-Contrastive Loss:}
The impact of using the proposed objective function (PCL) as the classifier’s loss function instead of the CE is examined. The performances of the proposed method with only the DFL, only the PCL and both embedded in the ResNet architecture are compared and the results shown in Table \ref{table:PCL_cifar}. It is clear that the results for only the DFL are better in the case of ’no defense’ training and, therefore, black-box attacks. On the other hand, using only the PCL improves the results in the case of adversarial training and, therefore, white-box attacks. Finally, combining both the components of the proposed method in one network achieves the required stability, convergence and better performances for all the adversarial attacks.

\begin{table}[h!]
\caption{Comparison between our proposed objective function (PCL) and the baseline approaches for CIFAR-10 dataset. \label{table:PCL_cifar}}
    \centering
    \begin{tabular}{|l|ccc|ccc|}
\hline 
\multicolumn{4}{|c|}{\textbf{No Defense}}\tabularnewline
\hline 
Defenses & FGSM & PGD-10 & C\&W \tabularnewline
\hline 
DFL & 44.60  & 35.40 & 48.50 \tabularnewline
\hline 
PCL & 42.50 & 31.40 & 43.10 \tabularnewline
\hline 
DFL + PCL & \textbf{43.85} & \textbf{38.70} & \textbf{53.07}\tabularnewline
\hline 
\multicolumn{4}{|c|}{\textbf{Adversarial Training}}\tabularnewline
\hline 
Defenses&  FGSM & PGD-10 & C\&W\tabularnewline
\hline 
DFL & 90.31  & 64.67 & 57.30\tabularnewline
\hline 
PCL& 90.61 & 70.45 & 58.20\tabularnewline
\hline 
DFL + PCL&  \textbf{91.80} & \textbf{74.50} & \textbf{61.30} \tabularnewline
\hline
\end{tabular}
\end{table}

\noindent\textbf{Effectiveness of using different backbones (WideResNet ) \cite{wide_resnet}:}
The performances of the proposed method using WideResNet architecture  as backbone are shown in Table \ref{table:PCL_cifar_wide}. Embedding the proposed method into WideResNet \cite{wide_resnet} improves the performance with reasonable margin that ranges from $0.8\%$ to $7.20\%$. This experiment shows the effectiveness of the proposed method with wide networks.

\begin{table}[h!]
\caption{Comparison between ResNet and WideResNet on CIFAR-10 dataset. \label{table:PCL_cifar_wide}}
    \centering
    \begin{tabular}{|l|ccc|ccc|}
\hline 
\multicolumn{4}{|c|}{\textbf{Adversarial Training}}\tabularnewline
\hline 
Backbone&  FGSM & PGD-10 & C\&W\tabularnewline
\hline 
ResNet&  91.80 & 74.50 & 61.30
\tabularnewline
\hline
WideResNet&  \textbf{92.60} & \textbf{77.20} & \textbf{68.51}
\tabularnewline
\hline
\end{tabular}
\end{table}

\begin{figure}
    \centering
      \includegraphics[width=0.9\linewidth]{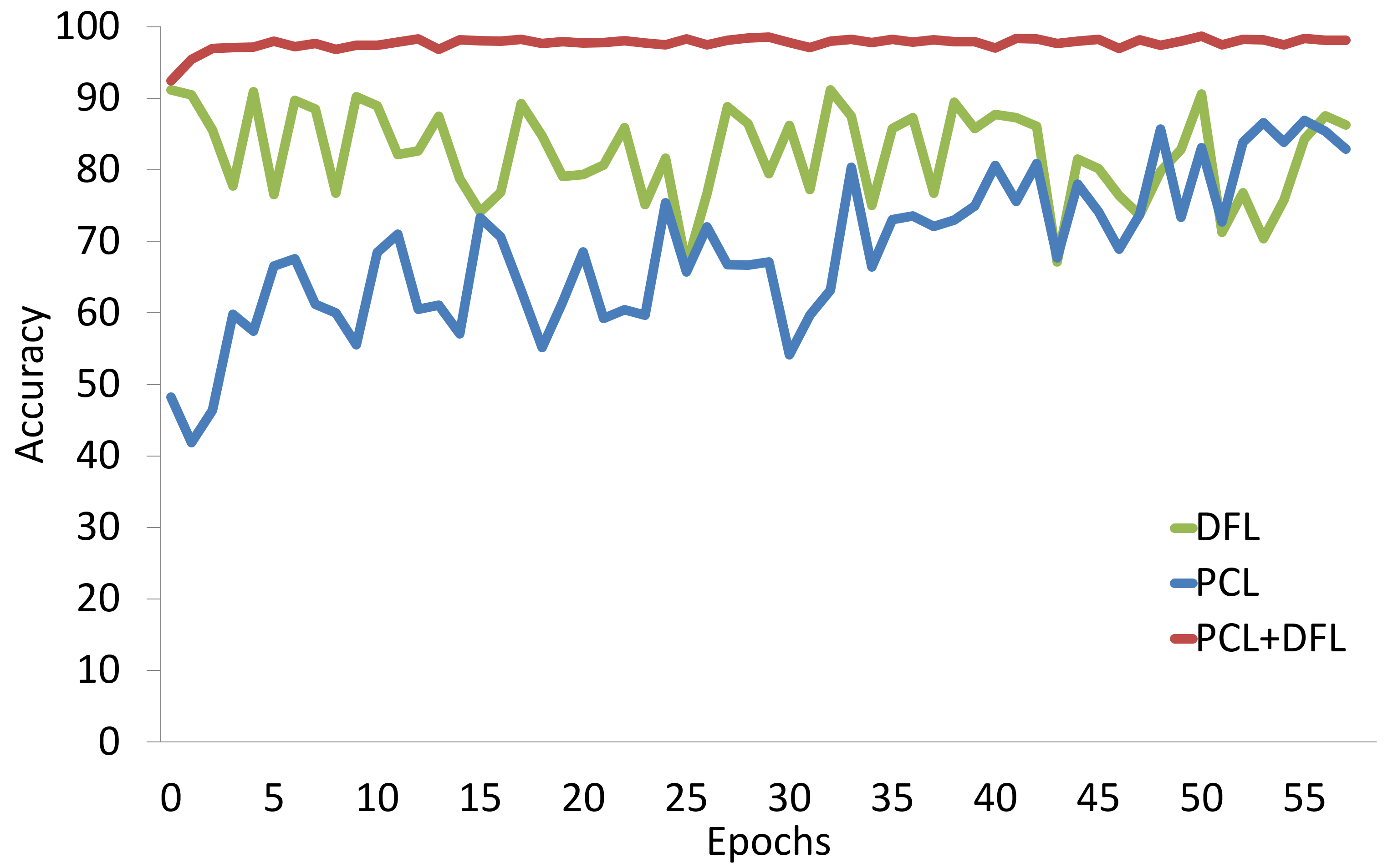}\\(a)\\
      \includegraphics[width=0.9\linewidth]{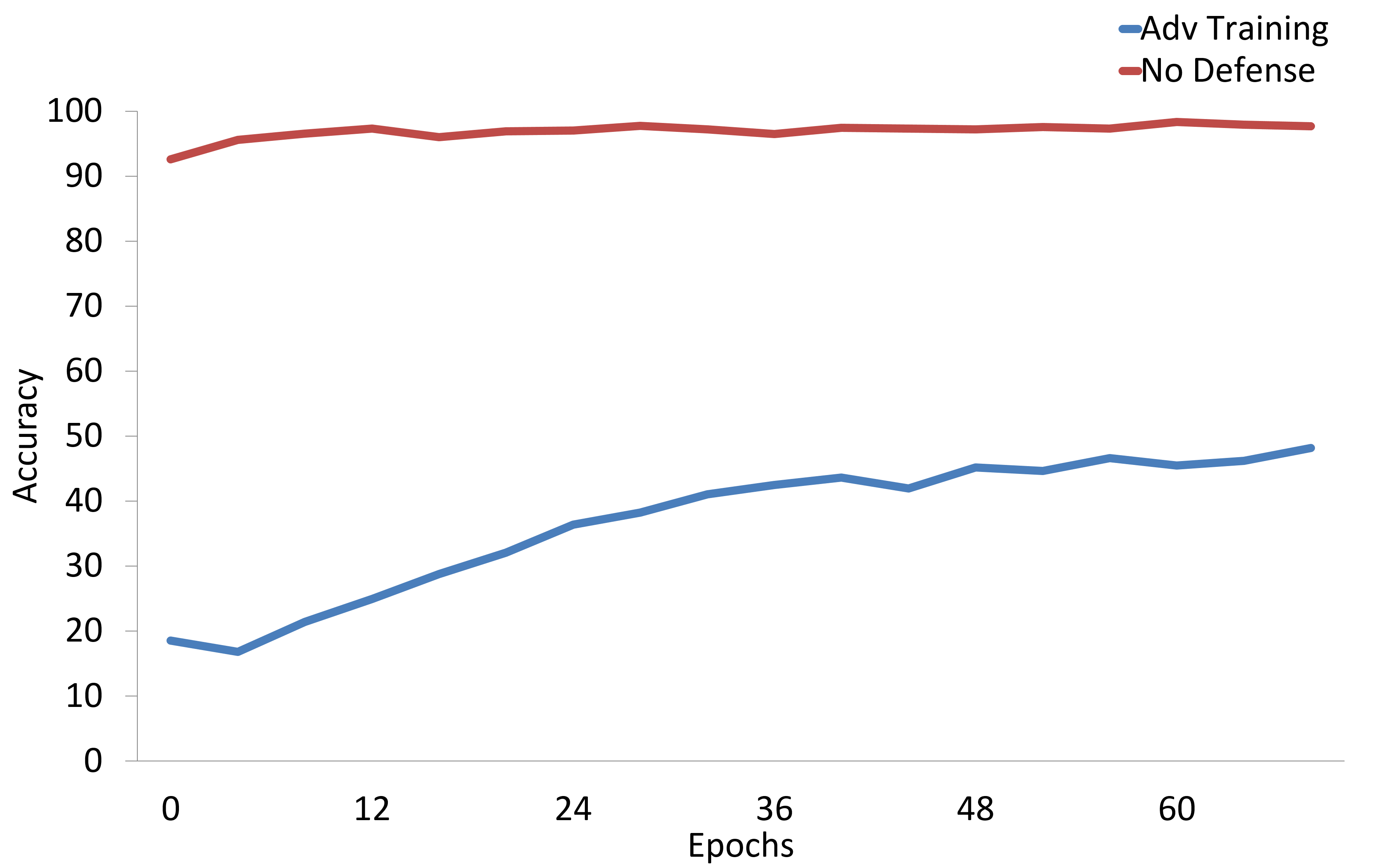}
      \\(b)   
   \caption{Comparison between a) the impact of DFL, PCL and both of them in the adversarial training, b) validation accuracy with 'no defense' and with adversarial training. }
  \label{fig:PCL_figures}
\end{figure}
The model’s validation accuracy values using the DFL, PCL and both together with the training conducted using a PGD as an attack are shown on the left-hand side in Figure \ref{fig:PCL_figures} (a). It can be seen that the PCL increases over time because it learns in a contrastive way while the DFL training process is almost steady which is a feature of the model regardless of any change in its input. However, combining both guarantees stable training and better performances. A comparison of the training/validation accuracy values with the ’no defense’ and adversarial settings using the MNIST dataset are shown in Figure \ref{fig:PCL_figures} (b).

\section{Conclusions}

In this paper, a generative adversarial defensive method for securing DNN models against data poisoning attacks is proposed. It is based on two types of defense that address the challenges of a network modification and gradient mask. The issues of degradation of accuracy and loss of convergence for very deep networks under the effects of adversarial attacks are identified. Then, a novel defense called the DFL is developed and it is shown through experiments that embedding it in a DNN is successful in solving these problems. An ablation analysis demonstrates that the DFL reduces the training error by $12\%$ and improves the accuracy by $7.1\%$ while quantitative comparisons prove that it represents the benchmarks for both ’no defense’ and adversarial settings. The very deep networks converge and learn better in the case of adversarial training when embedding the DFL. Also, we propose the PCL as an objective function to sharpen the decision boundaries amongst the classes in order to defend against white- and black-box attacks. Combining both these components demonstrates the robustness of the proposed method against different kinds of attacks, even with ’no defense’ training. The results illustrate that our approach is better than state-of-the-art ones by large margins of approximately $75\%$ and $55\%$ in the cases of ’no defense’ and adversarial training, respectively. This method highlights further necessary investigative directions regarding network modification techniques in terms of the adversarial paradigm.


\bibliographystyle{cas-model2-names}

\bibliography{ASOC}

\end{document}